\documentclass[]{statsoc}

\usepackage[a4paper]{geometry}
\usepackage{amsmath}
\usepackage{dsfont}
\usepackage{graphicx}
\usepackage{booktabs}
\usepackage{comment}
\usepackage{xcolor}
\usepackage{setspace}

\title[Learning High-Order Interactions via Targeted Pattern Search]{Learning High-Order Interactions via Targeted Pattern Search}
\author[Massi, Franco \textit{et al.}]{Michela C. Massi$^{1,2}$\footnote{Corresponding author: michelacarlotta.massi@polimi.it, Via Edoardo Bonardi, 9, 20133 Milan, Italy}, Nicola R. Franco$^{1}$, Francesca Ieva$^{1,2,3}$, Andrea Manzoni$^{1}$, Anna Maria Paganoni$^{1,2}$, Paolo Zunino$^{1}$}
\address{$^{1}$MOX, Department of Mathematics, Politecnico di Milano, Milan, Italy \newline
$^{2}$CADS-Center for Analysis, Decisions and Society, Human Technopole, Milan, Italy \newline
$^{3}$CHRP-National Center for Healthcare Research and Pharmacoepidemiology, University of Milano-Bicocca, Milan, Italy\newline}

\newcommand{\threshold}{supp_{\min}}

\begin{document}

\begin{abstract}
Logistic Regression (LR) is a widely used statistical method in empirical binary classification studies. 
However, real-life scenarios oftentimes share complexities that prevent from the use of the as-is LR model, and instead highlight the need to include high-order interactions to capture data variability.
This becomes even more challenging because of: (i) datasets growing wider, with more and more variables; (ii) studies being typically conducted in strongly imbalanced settings; (iii) samples going from very large to extremely small; (iv) the need of providing both predictive models and interpretable results.

In this paper we present a novel algorithm, \texttt{L}earning high-order \texttt{I}nteractions  via targeted \texttt{P}attern \texttt{S}earch (\texttt{LIPS}), to select interaction terms of varying order to include in a LR model for an imbalanced binary classification task when input data are categorical. \texttt{LIPS}'s rationale stems from the duality between item sets and categorical interactions. The algorithm relies on an interaction learning step based on a well-known frequent item set mining algorithm, and a novel dissimilarity-based interaction selection step that allows the user to specify the number of interactions to be included in the LR model. In addition, we particularize two variants (\texttt{Scores LIPS} and \texttt{Clusters LIPS}), that can address even more specific needs. Through a set of experiments we validate our algorithm and prove its wide applicability to real-life research scenarios, showing that it outperforms a benchmark state-of-the-art algorithm.
\end{abstract}


\section{Introduction}
Despite the evolution of statistical learning theory, Logistic Regression (LR) is still a commonly used statistical method in empirical studies in many research fields \citep{dreiseitl2002logistic}. It is widely regarded as the model of choice for situations where the occurrence of a dichotomous outcome is to be predicted from one or more covariates \citep{boateng2019review}. Examples can be found in medical research \citep{boateng2019review}, educational research \citep{niu2020review}, public health research \citep{lemon2003classification}, political sciences \citep{nicolau2007analysis}, economics \citep{sloane1994economics, zaghdoudi2013bank} and many others.

Regardless of the context of application, it is often the case that the logit of the expected value of the dichotomous response variable cannot be explained solely by additive functions of the predictors. In other terms, when the function $f(x_{1},x_{2})$ cannot be expressed as $g(x_{1})+h(x_{2})$ for some functions $g$ and $h$, we say that there is an interaction in $f$ between $x_{1}$ and $x_{2}$.

Many of the aforementioned fields of research share the need for introducing these interaction effects between their predictors to better infer on - and characterize - the outcome. For instance, in Genome-Wide Association Studies (GWAS) there is increasing awareness that \textit{epistasis}, or gene-gene interaction, plays a role in susceptibility to common human diseases \citep{moore2003ubiquitous, onay2006snp}. It has been argued \citep{moore2003ubiquitous} that epistasis is a ubiquitous component of the genetic architecture of common human diseases and that complex interactions are more important than the independent main effects of any susceptibility gene.

This calls for novel statistical approaches to identify the complex non-additive relationships between multiple variables to be included into a predictive model that would fully capture the underlying relationship with the target. However, when dealing with real-world applications of LR and interaction search, such as the one described above, several additional considerations must be taken into account.

First of all, nowadays datasets are growing \textit{wider}, with more and more variables, but the sample size may vary from extremely large to very limited. Linear models scale relatively well when handled via standard software, up to thousand of features together \citep{lim2015learning}. However, scalability becomes an issue when dealing with interactions.
Indeed, in a model with intercept and $p$ numerical predictors the total number of terms is given by

$$\sum_{k=0}^{p}\binom{p}{k} = 2^{p}.$$

This exponential growth gets even steeper in the presence of categorical covariates, as each one of the features' levels has to be considered. Nevertheless, dealing with a huge moltitude of categorical covariates is getting more frequent for instance in the aforementioned GWAS studies, where interactions between millions of Single Nucleotide Polimorphisms (SNPs) with triplets of levels affect even extremely rare mutations in the patients' phenotype \citep{llinares2019casmap, ceddia2020association}. This makes the inclusion of higher order interactions technically intractable, and causes the problem of finding interactions to fall in the framework of "large $p$ small $n$" problems, where the number of features significantly outweighs the number of observations and the \textit{curse of dimensionality} affects the reliability of statistical estimates.

In particular, when fitting LR models with no prior knowledge regarding the parameters, the traditional choice is Maximum Likelihood Estimation (MLE). However, a recent contribution from \cite{sur2019modern} shows how as $p$ grows w.r.t. $n$, estimates seem systematically biased, in the sense that effect magnitudes are overrestimated, they are far more variable than predicted using classical methods (such as Inverse Fisher Information), and inference measures, e.g. p-values, are unreliable especially at small values. This makes both prediction and interpretation of results extremely questionable and hinders the validity of the analysis.

When data exhibit complex dependencies, as well as high dimensionalities and small sample sizes, researchers are often also forced to tackle the issue of clustering observations in presence of class imbalance. This latter issue requires a further tailoring of the classification models that would otherwise incur in the risk of losing predictive power on the underrepresented class. This fact becomes a remarkable concern in application areas, such as
healthcare or insurance fraud, where identifying the underrepresented class correctly is
the most - if not the only - relevant matter.
\newline

\noindent For all these reasons, a new class of statistical methods needs to be designed in order 
to address the aforementioned complexities and provide LR models that can be effectively applied in most research domains.
Our present work tackles these needs and focuses on developing a novel approach to Interaction Learning specifically designed to: (i) identify a restricted number of the most relevant interactions of arbitrarily high order among \textit{categorical} features; (ii) optimize the selection of interactions for separating the classes in cases of strong imbalance; (iii) allow the user to keep the number of interactions to be included in the model under control, in order to foster significance and interpretability of the resulting LR. All these aspects were considered in the design of the proposed algorithm, while guaranteeing (iv) scalability, tractable computational times, and robustness to varying dimensionality, sample size and imbalance ratio.\\

\subsection{Interaction Learning Approaches and Our Contribution}
\label{related_works}
Discovering interactions is an active area of research \citep{lim2015learning}. However, when discussing related approaches in the multifaceted scenario our proposal deals with, it is important to make a distinction between two fairly independent lines of research that address different aspects of the subject matter. The first one deals specifically with categorical features - even though some studies try to expand their applicability to continuous features - and it is devoted to finding \textit{lists} of high-order feature interactions associated with the outcome.  One example can be found in \cite{shah2014random}. Research groups devoted to \textit{significant (discriminative) pattern mining} expand this concept by taking into account the statistical significance of the association, controlling the Family-Wise Error Rate (FWER), or the probability to detect false positive patterns \citep{llinares2015fast,papaxanthos2016finding, pellegrina2018efficient,sugiyama2019finding}. These works found a broad range of applications in some of the domains we mentioned beforehand, such as statistical genetics or healthcare \citep{llinares2019casmap, ceddia2020association}. However, their final objective is that of identifying all significantly associated interactions, without specific requirements on the number of selected patterns or their predictive power and tractability once fed to a classification model as LR.

Conversely, the other related research line works with both categorical and continuous features, and it focuses more on identifying the interactions to be used within non-linear models and Generalized Linear Models (GLM). Most of these methods deal with different regularization strategies to shrink models to the most useful primary effects and interaction terms  \citep{radchenko2010variable, rosasco2010regularization, bien2013lasso, lim2015learning}; a possible alternative falling into this class is to use a stochastic search algorithm for variable selection in Bayesian LR, as proposed by \cite{chen2011methods}. When selecting among very large numbers of features, one drawback of penalized models is the complexity of directly controlling for the number of terms introduced in the final classification model - despite their undeniable potential in producing powerful classifiers. Especially when dealing with high-order categorical interaction terms, even in the presence of strong penalizations, the actual number of terms in a LR might easily explode. As previously mentioned, the growing number of predictors may lead to an inflation of the estimates, hindering the attempt to make significant inference out of the resulting model, besides reducing the results interpretability.
\newline
\newline
\noindent Our contribution lies at the crossing of these two research lines. Indeed, we first develop a \textit{targeted} high-order interaction search algorithm which, from a set of categorical features, produces a list of useful interactions associated with the outcome and ranks them on the basis of their Odds Ratio (OR). Then, we introduce a novel feature selection method designed to pick from that list only a predefined number of interaction terms to be included in a LR model finally capable of well discriminating between the two classes (even in cases of small sample size and imbalanced setting) as well as yielding both interpretable results and reliable statistical inference.
The algorithms have been developed in Python 3 \citep{python}; the related package is available upon request.
\newline
\newline
\noindent This paper is structured as follows. In Section \ref{sec:method} we detail our methodology, starting from the main algorithm, \texttt{LIPS}, and then discussing a few possible alternatives. In Section 3 we validate our design choices for the proposed algorithms and provide empirical evidence that both \texttt{LIPS} and its variants are able to perform extremely well in binary classification tasks with categorical covariates, even under conditions of strong imbalance and low sample size.
In Section 4, by analyzing both simulated and real data, we compare our approaches with \texttt{glinternet} \citep{lim2015learning}, a state-of-the-art algorithm for interaction learning in contexts of LR. There we show that \texttt{LIPS} and its variants can perform comparably well, with the additional advantage of returning models that are much more interpretable.
In Section 5 we draw some conclusions and discuss possible future developments. Further algorithmic details and mathematical insights can be found in the Appendix.

\section{High-order interaction learning via targeted pattern search}
\label{sec:method}
In this section we formally introduce \texttt{LIPS}, the proposed \texttt{L}earning high-order \texttt{I}nteractions via targeted \texttt{P}attern \texttt{S}earch algorithm.

Note that the method has been developed to handle interactions among categorical covariates. Therefore, we will firstly provide some context and notation to describe the specific setting we are dealing with and its intrinsic peculiarities. Then, we will detail the two main steps of the algorithm, i.e. the preliminary \textit{targeted} pattern search and the subsequent novel dissimilarity-based interaction selection method, which together define the Interaction Learning core of \texttt{LIPS}.

Because of our interest in developing a methodology broadly applicable to most real-life research domains, we include in this work two variants of the main algorithm - namely \texttt{Scores LIPS} and \texttt{Clusters LIPS}. As further detailed in the last part of this section, these two alternative strategies are meant to satisfy specific user and data requirements, while also improving interpretability of the model.

\subsection{Notation}
Let us first provide some background and introduce the needed notation to follow the details of our methodology.
\newline
\noindent We are given $p$ categorical random variables $X_{1},\ldots,X_{p}$, where each $X_{j}$ takes values in some set of labels $\{l_{m_{1}}^{(j)},\ldots,l_{m_{j}}^{(j)}\}$; note that the number of levels $m_{j}$ depends on $j$, i.e. the categorical variables may have different supports.

First of all, we define the so-called dummy variables

$$X_{j}^{(i)} = \mathds{1}_{\{l_{m_{i}}^{(j)}\}}(X_{j})\quad\quad j=1,\ldots,p,\quad i=1,\ldots,m_{j},$$

\noindent so that $X_{j}^{(i)}$ is $\{0, 1\}$-valued and equals 1 if and only if $X_{j}$ attains the $i$th level.
Next, we introduce the collection of all possible interaction terms among the covariates $X_{1},\ldots,X_{p}$ as

$$\mathcal{I} = \left\{\prod_{j\in J}X_{j}^{(i_{j})}\quad\bigg{|}\quad J\subseteq\{1,\ldots,p\},\;i_{j}\in\{1,\ldots,m_{j}\}\right\},$$

\noindent so that each interaction is a binary r.v. that encodes a certain levels' combination. We adopt the standard convention by which the \textit{order} of an interaction corresponds to the number of terms therein minus one; for instance, $X_{1}^{0}X_{3}^{1}$ defines a first order interaction. With little abuse of notation we let $1\in\mathcal{I}$ be the empty interaction ($J = \emptyset$) and we allow for 0-order interactions, meaning that all dummies also fall in $\mathcal{I}$.

It is straightforward to see that
$$|\mathcal{I}|=(m_{1}+1)\cdot...\cdot(m_{p}+1),$$
indeed, when defining an interaction each $X_{k}$ leaves us with exactly $(m_{k}+1)$ choices, as we may either discard that variable or include one of its corresponding $m_{k}$ dummies. In particular, if $m_{1}=...=m_{p}=m$, then the cardinality of $|\mathcal{I}|=(m+1)^{p}$ grows exponentially with $p$.

\medskip

For more readability, let us introduce a few more preliminary definitions:
\begin{enumerate}
\item For an interaction $T\in\mathcal{I}$ we write $|T|$ to denote the number of dummy variables involved in the product representation of $T$;
\item Given two interactions $T,\;S\in\mathcal{I}$ we say that $T$ is a subinteraction of $S$ if all its dummies appear in the product expansion of $S$; equivalently, $S = T\cdot Z$ for some $Z\in\mathcal{I}$;
\item For $T, S\in\mathcal{I}$ we define their maximum common divisor, MCD($T$, $S$), as the highest order interaction $Z\in\mathcal{I}$ that is both a subinteraction of $T$ and $S$;
\item We say that two interactions $T$ and $S$ are incompatible, and we write $T\perp S$, if they respectively include two different levels of the same random variable and therefore $T\cdot S$ is identically zero.
\end{enumerate}

\subsection{Interaction Learning in the \texttt{LIPS} algorithm}
\label{sec:LIPS}
We are given a dataset $\mathcal{D}=\{(x_{1,i},\ldots, x_{p,i}, y_{i})\}_{i=1}^{n}$ consisting of $n$ i.i.d. observations of $p$ categorical covariates $X_{1},\ldots, X_{p}$ and a binary outcome $Y$.
We assume to be in an imbalanced setting with respect to $Y$, that is we suppose the classes $\mathcal{O}=\{i\;|\;y_{i}=1\}$ and $\mathcal{Z}=\{i\;|\;y_{i}=0\}$ to come in remarkably different proportions. Without loss of generality, we consider the case of $\mathcal{O}$ being the minority class, which reflects the situation of $\mathds{P}(Y=1)\ll \mathds{P}(Y=0)$.

\smallskip

Our first purpose is to deduce from the data a suitable subcollection of interactions that is informative w.r.t. $Y$ but small in size. To achieve such goal, we perform two steps: identification of all the candidates interactions through a targeted pattern search; choosing of the top $K$ most relevant interactions via our novel dissimilarity-based interaction selection method, where $K$ is user-specified and tipically satifies $K\ll|\mathcal{I}|$.

\subsubsection{Targeted pattern search}
\label{targeted_ps}
In principle, starting from the data in $\mathcal{D}$, it is very easy to compute the sample values of any given $T\in\mathcal{I}$. However, as in general the cardinality of $\mathcal{I}$ is huge, it is seemingly impossible to study each conceivable interaction term alone; it is this very drawback that whence the need of finding a preliminary subset of candidates $\hat{\mathcal{I}}\subset\mathcal{I}$.

To address this task, we propose to focus only on those interactions that are different from zero within the minority class $\mathcal{O}$ with an empirical frequency that is higher than a fixed threshold $\threshold>0$.

This choice, which we further discuss in Section \ref{sec:OCL&diss-based-sel} and we will also refer to as \textit{One-Class-Learning} (OCL), is mainly driven by the need of undertaking the imbalance between the two classes, but has also a few advantages in terms of computational cost.

In what follows, for an interaction $T\in\mathcal{I}$ we denote by $t_{i}$ its value corresponding to the $i$th statistical unit in the dataset. As mentioned above, we define

$$\hat{\mathcal{I}} := \left\{T\in\mathcal{I} \;\text{ such that }\;\frac{1}{|\mathcal{O}|}\sum_{i\in\mathcal{O}}t_{i}>\delta\right\}.$$

Since it is not feasible to compute such frequencies for all $T \in \mathcal{I}$, to determine $\hat{\mathcal{I}}$ we can rely on the duality between the concepts of interactions and \textit{itemsets}. The latter, which are also referred to as \textit{patterns}, are tipically found in problems of association rule learning such as market basket analysis \citep{freqitemsets}, intrusion detection \citep{intrusion} and others. There, one is given a set of possible \textit{items} and a list of \textit{transactions}, which are nothing but sets of items. The goal is then to mine the list of such transactions in order to find which subsets of items (itemsets) occur with a sufficient frequency (also called \textit{support}).

In our framework, we observe that there is a one-to-one correspondence between itemsets and interactions, as these can be thought as being itemsets of dummy variables. In fact, if we interpret dummy variables as items, then the observations $\{(x^{(1)}_{1,i},\ldots,x^{(m_{p})}_{p,i})\}_{i=1}^{n}$ are nothing but transactions (where an item is picked if and only if its corresponding dummy equals 1); similarly, a certain pattern of dummies is present in the transaction if and only if the corresponding interaction term is nonzero.

This property is very useful as it allows us to take advantage of all the wide variety of frequent item set mining algorithms, which are recently becoming more and more efficient \citep{freqitemsets, shah2014random}. In particular, we may reformulate the problem of computing $\hat{\mathcal{I}}$ as that of finding a list of patterns that appear with a frequency greater than $\threshold$ (minimum support) within the minority class.
Many possible implementations - such as the \textit{Apriori} algorithm \citep{apriori1994}, which we will later make use of - are available for our purpose, and it can be convenient to test different ones depending on the data.

\smallskip

After having constructed the candidate set $\hat{\mathcal{I}}$, preliminary to the next phase we build for each $T\in\hat{\mathcal{I}}$ the contingency table of $(T, Y)$ along the whole dataset $\mathcal{D}$. This task requires, at least, to compute the corresponding frequencies in the majority class, $\frac{1}{|\mathcal{Z}|}\sum_{i\in\mathcal{Z}}t_{i}$. We point out that, besides the possibly high-cardinality of $\hat{\mathcal{I}}$, this step can now be performed very efficiently: for more technical details, refer to the Appendix, Section A.1.

\subsubsection{A dissimilarity measure based feature selection algorithm}
\label{sec:dissimilarity}
Knowing the contingency table $(T, Y)$ for every $T\in\hat{\mathcal{I}}$ allows us to associate to each candidate interaction an odds-ratio:

$$\text{OR}_{T}:=\frac{\left(\sum_{i:\;t_{i}=1}y_{i}\right)/\left(\sum_{i:\;t_{i}=1}(1-y_{i})\right)}
{\left(\sum_{i:\;t_{i}=0}y_{i}\right)/\left(\sum_{i:\;t_{i}=0}(1-y_{i})\right)}.$$ 

We use this statistic to rank the candidate interactions in $\hat{\mathcal{I}}$: more precisely, we sort the patterns in descending order according to the quantity $\log|\text{OR}_{T}|$, so that reciprocal odds-ratios are treated equally.

Next, given a fixed $K\in\mathds{N}$, we propose a way to extract from the sorted list of candidates $\hat{\mathcal{I}}$ a subset of (at most) $K$ interactions that are suitable to perform inference on $Y$. Naively, one may think of selecting the first $K$ patterns in the list. However, this approach may present several drawbacks (cfr. Section \ref{sec:OCL&diss-based-sel}): due to the combinatorial nature of high-order interactions, such strategy could easily result in a list of nested patterns (subinteractions), thus carrying redundant information; if $K$ is small, this may lead to overfitting-like phenomena, as the feature space is not explored sufficiently.

\smallskip

To overcome this drawback, we introduce a \text{dissimilarity measure}, $d:\mathcal{I}\times\mathcal{I}\to[0, +\infty)$, defined as follows

\begin{equation}
\label{eq:dissimilarity_measure}
d(T, S):=\begin{cases}
|T|\vee|S| & T\perp S\\
|T|\vee|S| - |\text{MCD}(T, S)| & \text{otherwise}
\end{cases},
\end{equation}

 \noindent where $x\vee y:=\max{\{x, y\}}$. This allows us now to compare two different patterns.

\smallskip

By definition, $d$ returns higher dissimilarities for patterns that involve completely different variables and for those that are incompatible. Because of these two properties, we may consider $d$ a suitable measure for exploring different regions of the feature space; see Section B in the Appendix for further insights about the definition of $d$.

\smallskip
\noindent By making use of the dissimilarity measure $d$, we extract $K$ patterns from the sorted list $\hat{\mathcal{I}}$ according to the following iterative procedure:
\begin{itemize}
    \item[1.] We remove from $\hat{\mathcal{I}}$ the first interaction $T$, which corresponds to the one that maximizes $\log{|\text{OR}_{T}|}$, and place it in a new list $\mathcal{I}_{K}$;
    \item[2.] We search for those interactions in $\hat{\mathcal{I}}$ that are the most dissimilar from the ones in $\mathcal{I}_{K}$. In other words we determine argmin$_{T\in\hat{\mathcal{I}}}\min_{S\in\mathcal{I}_{K}}d(T,S)$. Among these, we select the one with highest rank and move it from $\hat{\mathcal{I}}$ to $\mathcal{I}_{K}$;
    \item[3.] We repeat the instructions in (2) until $\mathcal{I}_{K}$ contains $K$ interactions.
\end{itemize}

In the end, we are left with a collection $\mathcal{I}_{K}=\{T_{k}\}_{k=1}^{K}$ of $K$ interactions which can now be used as predictors in a logistic model with response $Y$. In other words, \texttt{LIPS}'s resulting model reads

\begin{equation}
    \label{LIPS}
    \text{logit}{\mathds{P}}\left(Y=1|T_{1},\ldots,T_{K}\right) = \beta_{0}+\sum_{k=1}^{K}\beta_{k}T_{k}
    \text{ .}
\end{equation}

\noindent Especially when the sample size is not large enough, it can be convenient to primarily filter the candidate list $\hat{\mathcal{I}}$ by removing those interactions that have an odds-ratio suspiciously close to 1. For example, as we did in our case study (cfr. Section \ref{sec:second_example}), one may choose $\gamma\in(0,1)$ and discard all those patterns whose $\gamma$-level confidence interval for the odds-ratio contains the value 1.
In general, this will not affect the ranking and the final steps described above, but may result in more reliable models.

\subsection{Variants}
\label{sec:variants}
In this paragraph we explore two variants of our algorithm which aim at increasing interpretability and tractability of the final model. 
Although in our context of categorical predictors even high-order interactions remain highly interpretable (as aforementioned, they encode the co-presence of multiple levels of different variables), further reductions in the model structure can be of great interest in certain applications (cfr. Section \ref{sec:discussion}).

Below we discuss two possible approaches: \texttt{Scores LIPS}, which condenses all the information into just a pair of variables; \texttt{Clusters LIPS}, that is instead a possible compromise between the two, where the $K$ identified interactions are grouped into multiple \textit{Compatibility Clusters}.
Note that, in both cases the number of terms in the model is smaller than or at most equal to $K$. This makes the two alternatives a go-to solution in case we needed to include the information carried by several interactions (for instance, in a $p \gg n$ setting), without incurring in the risk of ending with an LR model that has too many terms w.r.t $n$.

As both variants require a further reshaping of the information carried by the selected interaction terms, we can expect their performances to be slightly worse w.r.t. the base \texttt{LIPS} where all terms are included independently.
However, as mentioned beforehand - and further expanded in the Discussion (Section \ref{sec:discussion}) - these variants present several advantages in terms of both interpretability and dimensionality reduction, two aspects that play a key role in certain research scenarios and make these algorithms widely applicable.

\subsubsection{Risk and Protection Scores: \texttt{Scores LIPS}}
\label{sec:scores}
In Section 2.2, we ranked the candidate interactions in a unique list using the log odds-ratio as criteria. However, we may avoid mixing together terms that exert opposite influences on the target and instead split the collection $\hat{\mathcal{I}}$ into two sublists

$$\hat{\mathcal{R}}:=\{T\in\hat{\mathcal{I}}\;|\;\text{OR}_{T}>1\},
\quad
\hat{\mathcal{P}}:=\{T\in\hat{\mathcal{I}}\;|\;\text{OR}_{T}<1\},$$

which we may now sort separately, respectively by descending and ascending odds-ratios. We refer to the two collections as \textit{risk} and \textit{protection patterns}, a terminology that is mostly motivated by the majority of clinical applications.

Given an even integer $K$, we then apply our dissimilarity selection algorithm in order to extract two groups, $\hat{\mathcal{R}}_{K/2}\subset\hat{\mathcal{R}}$ and $\hat{\mathcal{P}}_{K/2}\subset\hat{\mathcal{P}}$, each of dimension $K/2$. Then, we construct the Risk Score $R$ and Protection Score $P$ as

$$R:=\sum_{T\in\hat{\mathcal{R}}_{K/2}}T,
\quad\quad\quad
P:=\sum_{S\in\hat{\mathcal{P}}_{K/2}}S.
$$

\noindent The intuition between the two scores is that $R$ (resp. $P$) counts the number of selected risk (protection) patterns that are present within a certain observation. In this way, the information relevant for inferring on $Y$ gets squished into a single pair of highly interpretable variables. The final model is then

\begin{equation}
    \label{scores}
    \text{logit}{\mathds{P}}\left(Y=1|R, P\right) = \mu_{0}+\alpha R+\beta P.
\end{equation}

\subsubsection{Compatibility Clusters: \texttt{Clusters LIPS}}
\label{sec:clusters}
The approach described in the previous section has the advantage of substituting the $p$ original covariates $X_{i}$ with two scores that carry information about possible interactions. However, compressing the two lists, $\hat{\mathcal{R}}_{K/2}$ and $\hat{\mathcal{P}}_{K/2}$, directly into the scores $R$ and $P$ could be too rough.

Here we propose milder aggregation criteria, the \textit{compatibility clusters}, that are based on the idea that we should not bind together incompatible patterns. In principle, in fact, no observation can ever present two incompatible patterns simultaneously; therefore, we may want to weight them differently in our model.
Concretely, after having defined $\hat{\mathcal{R}}_{K/2}$ and $\hat{\mathcal{P}}_{K/2}$ as in the previous section, we focus on each of them separately and propose a way to aggregate their terms. Starting from $\hat{\mathcal{R}}_{K/2}$, we wish to find a partition $\hat{\mathcal{R}}_{K/2}^{1}\cup...\cup\hat{\mathcal{R}}_{K/2}^{a}=\hat{\mathcal{R}}_{K/2}$ for which: 1) all interactions within the same subgroup $\hat{\mathcal{R}}_{K/2}^{j}$ are compatible with one another; 2) given two subgroups there always exists at least a pairing of incompatible patterns, that is: no subgroups can be fused together without violating the first rule. 

Stated as above, there is no unique way of doing this. In our experiments, we decided to construct the groups s.t. their total number was as small as possible, thus maximazing the information compression. Further details on the implementation can be found in the Appendix, Section A.2.

\smallskip

Performing the described steps also for the protection patterns allows us to define the cluster scores

$$R_{j}=\sum_{T\in\hat{\mathcal{R}}_{K/2}^{j}}T,
\quad
P_{j}=\sum_{T\in\hat{\mathcal{P}}_{K/2}^{i}}T,$$

for $j=1,\ldots,a$ and $i=1,\ldots,b$. The final model reads:

\begin{equation}
    \label{clusters}
    \text{logit}{\mathds{P}}\left(Y=1|R^{1},\ldots,R^{a}, P^{1},\ldots,P^{b}\right) = \mu_{0}+\sum_{j=1}^{a}\alpha_{j}R_{j}+\sum_{i=1}^{b}\beta_{i}P_{i}\text{ .}
\end{equation}

\noindent Comparing \eqref{clusters} with \eqref{scores}, we see that now we are splitting the risk and protection scores into more sub-scores, arguing that each one describes different, possibly incompatible, situations.\\
\\

\section{Simulation Study}
This section is devoted to verify some of the statements made on the most relevant aspects of the proposed algorithm throughout the paper.
In order to provide these empirical evidences in a controlled setting, we had to build a suitable simulation protocol. Indeed, to be meaningful, we had to construct a dataset where the binary outcome showed complex dependencies with multiple categorical covariates simultaneously, so that learning the right interactions would be essential to separate the two classes. In Section \ref{sim_dataset} we describe how the needed data was defined and simulated; then, on such data, we compare the performances of \texttt{LIPS} and its variants with those achieved by other logistic models that do not account for interaction effects.

The subsequent experiments are instead devoted to verifying whether the \textit{"targeted" search} or One Class Learning (OCL), focused on identifying patterns in the positive class only, could actually compete against the same algorithm fed with lists of interactions extracted from both classes. Then, we validate the value added by the \textit{dissimilarity-based} feature selection in building the final set of interactions to include in the LR, w.r.t. picking the top $K$ interactions ranked on $log|OR_{T}|$. Finally, we verify the robustness of \texttt{LIPS} to variations in sample size and class imbalance rate.

\subsection{Simulated Data}
\label{sim_dataset}

\begin{figure}[b]
\centering
\makebox{\includegraphics[width=\columnwidth]{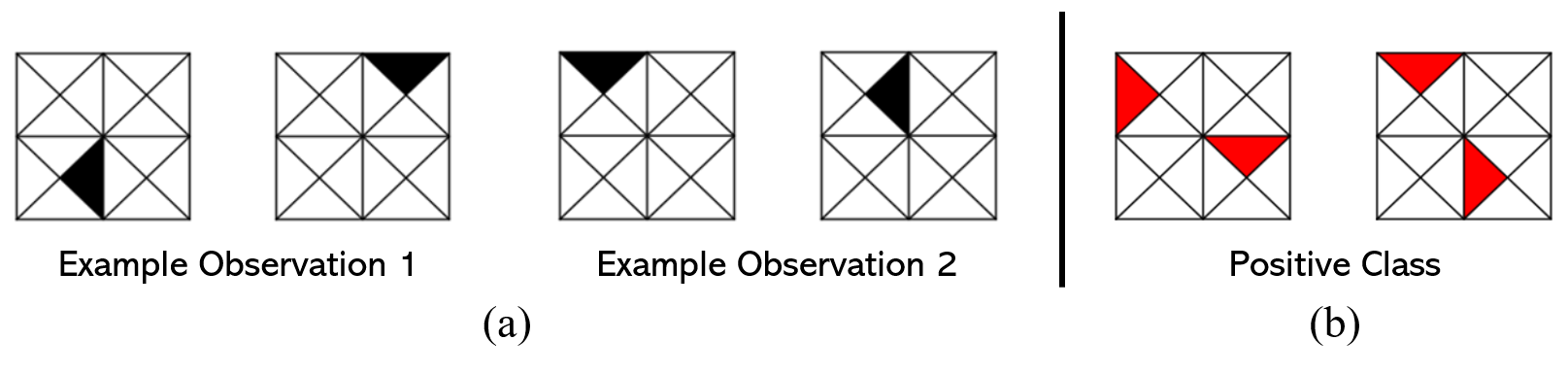}}
\caption{Geometric model behind simulated data. On the left (a) are two illustrative observations in the dataset, represented by two tiles from each squared tiling. On the right (b) represents the tiles that determine the positive class.}
\label{fig:simdata}
\end{figure}

For the definition of the datasets adopted in all the simulation experiments described in the following, we exploited a geometric model. This allowed us to synthetically introduce the aforementioned outcome-covariates dependency based on \textit{multiple interactions}.
\newline
\newline
The underlying geometric model is defined as a double squared tiling structure (Fig.\ref{fig:simdata}). Each instance in the dataset is represented by a couple of triangular tiles, one picked from the left tiling and one from the right with uniform probability. Two example instances are reported in Fig.\ref{fig:simdata}.a.
All the covariates in the dataset are binary representations of the position of one of the two tiles (from left or right tiling). Indeed, the position of the first tile on the left is described by 5 binary features defined as the following:
\begin{enumerate}
    \item $R_{1}$ (\textit{right}) takes value 1 if the tile is in the right half, 0 otherwise.
    \item $U_{1}$ (\textit{up}) takes value 1 if the tile is located in the upper half of the tiling, 0 otherwise.
    \item $D_{1}$ (\textit{diagonal}) takes value 1 if the tile is located below the main diagonal of the tiling, 0 otherwise.
    \item $A_{1}$ (\textit{anti-diagonal}) takes value 1 if the tile is located below the anti-diagonal, 0 otherwise.
    \item $O_{1}$ (\textit{outside}) takes value 1 if the tile is outside the \textit{inner square} (the 45-degrees rotated square - composed of 8 triangular tiles - that has its vertices on the mean points of the tiling's edges), 0 otherwise.
\end{enumerate}
Similarly, $R_{2}$, $U_{2}$, $D_{2}$, $A_{2}$ and $O_{2}$ describe the position of the second tile in the right squared tiling. Notice that to recover the notation adopted in Section \ref{sec:method} one may simply let $X_{1}:=R_{1},\; X_{2}:=U_{1},\dots,\;X_{10}:=O_{2}$.  

The dependent variable $Y$, that defines the two classes ($y=\{0,1\}$), signals whether at least one of the two tiles of the observations falls within one of the red areas highlighted in Fig. \ref{fig:simdata}.b. As such, $Y$ is univocally determined by the 10 covariates described above. However, in order to work in a non-deterministic, setting, we artificially added noise to the data by mislabeling some observations. In particular:
\begin{enumerate}
    \item If both tiles of instance $i$ (left and right) fell in one \textit{red area} (hence, $y_{i}=1$), the label was changed to $y_{i}=0$ with probability $\tilde{p}=0.005$.
    \item In all other cases, the value of $y$ was changed to the opposite class with a probability $\tilde{q}=0.05$.
\end{enumerate}
There are several reasons motivating this construction of the dataset. First of all, note that the two classes are unbalanced by construction. Indeed, before introducing the noise, one has $\mathds{P}(Y=1) = 1-\mathds{P}(Y=0) = 1-\left(\frac{14}{16}\right)^{2} \simeq 23.4\% $, as the two tiles are extracted independently and, due to uniform distribution, each one of them has a probability of 14/16 of not falling within a red zone.
Secondly, strong dependencies exist within groups of covariates: for instance, if $R_{1,i}=U_{1,i}=1$, then necessarily $A_{1,i}=1$. Lastly, the minority class ($y_{i}=1$) is built in such a way that attaining a good classification performance would be impossible without introducing interaction terms, as the red tiles need to be described with more variables simultaneously to be properly identified in the tilings.

\subsection{Learning with or without interactions}
\label{sec:first_example}
As mentioned, the simulated data were built with the aim of challenging the LR models that try to classify observations exploiting the additive effect of the predictors alone. 
As a starting point, we hence compare the performance of a LR and a Lasso Regression without interaction terms against our proposed algorithm in its three variants.
For this experiment, we adopted the previously described procedure in order to generate 10 datasets, each of $n=$10,000 observations with $\mathds{P}(Y=1)\simeq 0.234$. We then splitted each dataset into training and test sets according to a 70/30 ratio. In Table \ref{tab:firstEx} we reported the following metrics: Area Under the ROC Curve (AUC), Sensitivity, Specificity, Negative Predictive Value (NPV) and Positive Predictive Value (PPV).
It is clear from Table \ref{tab:firstEx} that considering only main effects and their additive contribution makes the model practically useless in discriminating between the two classes ($AUC=0.49\pm0.015$ for both LR and Lasso). The Lasso Regression was performed imposing $\lambda=10^{-5}$, because larger $\lambda$ values shrinked the model to no term at all.

\begin{figure}
\centering
\makebox{\includegraphics[width=\columnwidth]{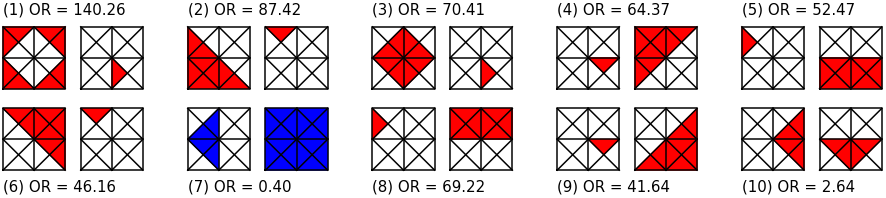}}
\caption{$K$ patterns identified by \texttt{LIPS} in one trial of the first simulation experiment (where $K=10$). Tiles are colored according to the areas defined by the categorical terms involved in each of the selected interactions. Red patterns are considered \textit{risk patterns} ($OR>1$), while \textit{protection patterns} ($OR<1$) are colored in blue.}
\label{fig:LIPS_patterns}
\end{figure}

Note that our algorithm (here with $K = 10$ and $\threshold = 10\%$) is yielding a very high score in all its variants. Nonetheless, it is not surprising to see \texttt{LIPS} using as-is interactions obtain the highest AUC among the three.
\newline
\newline
\noindent A remarkably positive aspect about building statistical models with high-order interactions among categorical features, is the undeniable interpretability of the resulting model itself. This is especially true when applying \texttt{LIPS}, that returns an arbitrarily long list of high-order interaction terms selected in order to (i) be highly descriptive of one class in particular and (ii) cover as much diverse information as possible regarding such class - thanks to the dissimilarity-based selection.
In Figure \ref{fig:LIPS_patterns} we provide a concrete demonstration of this claim. We represented graphically the 10 interaction terms included in the model generated by \texttt{LIPS} in one of the trials of this experiment. The tiles in each pair of squared tilings are colored higlighting the areas defined by the dummies included in each of the $K=10$ selected interactions. Red and blue coloring depend on the OR. The order in which the terms are drawn (to be read by row), is the same order in which the algorithm picked each of them. This means, for instance, that the only blue pattern was selected as $6^{th}$ interaction to be included. It can be easily noticed how each of the selected interaction describes precisely certain aspects of the two classes. For instance, the first pattern spots one triangle in the right tiling that appears to be associated with the minority class, which is coherent with the original construction of the data (cfr. Figure \ref{fig:simdata}.b). In contrast, the second selected interaction highlights a completely different (but still consistent) area. It is interesting to see how the only \textit{protection pattern}, despite not telling much about the right tiling, precisely defines an area in the left one where no minority class observation should fall.

\begin{table}
\caption{\label{tab:firstEx}Results on the simulated dataset for LR, Lasso Regression and \texttt{LIPS} in its three variants. The columns report model performances in terms of Area Under the ROC Curve (AUC), Sensitivity, Specificity, Negative Predictive Value (NPV) and Positive Predictive Value (PPV).}
\resizebox{\textwidth}{!}{
\fbox{%
\begin{tabular}{*{7}{l}}
\textbf{Model} & \textbf{Terms} & \textbf{AUC} & \textbf{Sensitivity} & \textbf{Specificity} & \textbf{NPV} & \textbf{PPV} \\ \hline
LR & 20 $\pm$ 0.000 & 0.494 $\pm$ 0.015 & 0.465 $\pm$ 0.069 & 0.603 $\pm$ 0.105 & 0.299 $\pm$ 0.035 & 0.759 $\pm$ 0.015\\
Lasso & 9.900 $\pm$ 0.316 & 0.494 $\pm$ 0.015 & 0.464 $\pm$ 0.068 & 0.604 $\pm$ 0.103 & 0.299 $\pm$ 0.034 & 0.760 $\pm$ 0.015\\
LIPS & 10 & \textbf{0.916 $\pm$ 0.010} & \textbf{0.854 $\pm$ 0.016} & \textbf{0.949 $\pm$ 0.038} & \textbf{0.948 $\pm$ 0.008} & \textbf{0.865 $\pm$ 0.088}\\
Scores LIPS & \textbf{2} & 0.831 $\pm$ 0.026 & 0.704 $\pm$ 0.060 & 0.845 $\pm$ 0.070 & 0.890 $\pm$ 0.019 & 0.631 $\pm$ 0.088\\
Clusters LIPS & 7.300 $\pm$ 0.483 & 0.842 $\pm$ 0.032 & 0.704 $\pm$ 0.060 & 0.845 $\pm$ 0.070 & 0.890 $\pm$ 0.019 & 0.631 $\pm$ 0.088\\
\end{tabular}}}
\end{table}

\subsection{One-Class Targeted Search and the relevance of Dissimilarity-based Interaction Selection}
\label{sec:OCL&diss-based-sel}
\begin{figure}
\centering
\makebox{\includegraphics[width=0.9\columnwidth]{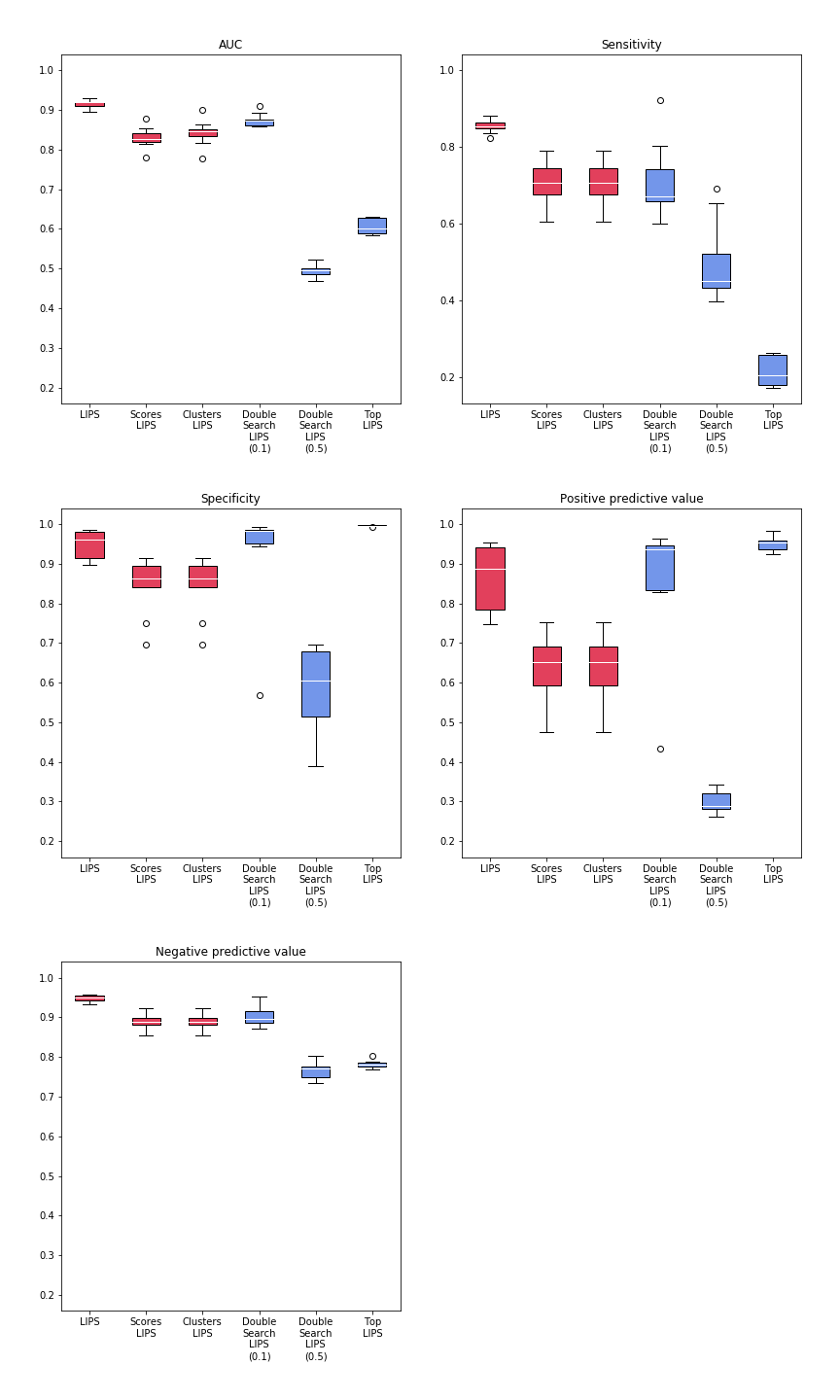}}
\caption{In order, performance of \texttt{LIPS} in its three variants (red), against DS-\texttt{LIPS} with $supp_{min}=0.1$, DS-\texttt{LIPS} with $supp_{min}=0.5$ and TOP \texttt{LIPS} (blue) on simulated data.}
\label{fig:LIPSvsvariants}
\end{figure}

Because of its combinatorial character, the search for high-order interactions remains a computational challenge, irrespectively of whether the objective is to fit a classification model or to identify all relevant patterns. As extensively discussed above, \texttt{LIPS} algorithm exploits the \textit{interaction-item set} duality to extract relevant high-order interactions from the training data. To do that, it employs the widely recognized Apriori algorithm for frequent item set mining, focusing the search on the positive class observations only.
Among other things, the experiment conducted in our simulated setting is meant to support this approach by empirically demonstrating how it does not hinder the classification of both classes, and it may even foster performance while saving time w.r.t. extracting item sets from the two. 
\newline
\newline
Once proven the validity of this first relevant aspect of the proposed algorithm, the second fundamental design choice that deserves an empirical evaluation is the dissimilarity-based feature selection method. Indeed, this peculiar passage builds arbitrarily long lists of features (interactions) by selecting the most diverse among those with the highest OR. This method was designed under the assumption that a more diverse set of interactions might reduce noise, generalize better on unseen minority class observations by capturing the whole underlying class' distribution and collect the most useful interactions to interpret and characterize this group. We tested these ideas within the same experiment as above.

We exploited the same 10 datasets described in Section \ref{sec:first_example} but trained four different algorithms: 
\begin{enumerate}
    \item \texttt{LIPS} in its characteristic \textit{One-Class-Learning} (OCL) version, searching on minority class examples only with $supp_{min} = 0.1$. We tested the algorithm in all its three variants (\texttt{LIPS}, \texttt{LIPS} Scores and \texttt{LIPS} Clusters).
    \item \textit{Two-Class-Learning} \texttt{LIPS} (referred to as Double Search \texttt{LIPS}), building the list of interactions searching separately in both classes. Similarly to OCL \texttt{LIPS} we imposed $supp_{min}=0.1$.
    \item Double Search \texttt{LIPS} with $supp_{min}=0.5$.
    \item To evaluate the substantiation of the hypotheses on the value added by the dissimilarity-based feature selection, the fourth algorithm is a version of the OCL \texttt{LIPS} that only picks the top $K$ interaction terms from the ranked list (Top \texttt{LIPS}).
\end{enumerate}
In Figure \ref{fig:LIPSvsvariants} we provide the results of this experiment.
The most traditional implementation of \texttt{LIPS} outperforms all other versions basically on all the considered metrics. First of all, note that the DS-\texttt{LIPS} with the same $supp_{min}$ performs almost comparably in terms of AUC, but Sensitivity is strongly affected by searching for patterns within both classes. This result supports our OCL approach to foster classification accuracy specifically on the underrepresented class.

Moreover, the DS-\texttt{LIPS} with the highest support performs worst on all dimensions: at first sight, this is surprising as one would expect highly frequent patterns in both classes to separate the two at best. However, raising the $supp_{min}$ value, especially for what concerns the minority class, can result in a poor performance whenever the separating patterns are very different and scattered along the data.

For what concerns \texttt{Top LIPS}, its performance is particularly low in terms of Sensitivity. Comparing this metric with the others, it is straightforward to deduce that the algorithm's performance is hindered by a very high number of False Negatives ($Sensitivity \simeq 0.2$).

Interestingly, this latter result highlights a possible reason why the dissimilarity-based
interaction selection step may enhance the identification of the real underlying distribution of minority class observations in a generalized and robust manner. Indeed, the $K$ interactions selected by \texttt{Top LIPS} with their high OR are very specific to certain minority class observations ($PPV \simeq 1$), thus guaranteeing a precise majority class classification as well ($Specificity \simeq 1$, as the identified patterns are extremely rare in the overrepresented class), but the chance to generalize them to the whole minority class is quite low. In other words, the algorithm is including a set of interactions that exist within the positive class only, however they all similarly describe one \textit{subgroup} of this class - which appears to be quite distant from all other examples in the dataset. This situation may arise in presence of outliers within the underrepresented class, or in case of an irregularly distributed minority class with two or more subgroups of observations. As a matter of fact, note that the simulated data has evident subgroups, as positive observations are defined by several distinct areas in the geometric space.
This testifies in favour of the robustness of the dissimilarity-based feature selection method that would, by design, include at least one of the high OR patterns (thus describing the subgroup), but would then be forced to add \textit{dissimilar} interaction terms, lowering the risk to overfit the group of outliers, and providing a better generalization on the whole minority class population. 

\subsection{On the robustness to Sample Size}
\begin{figure}[t]
\centering
\makebox{\includegraphics[width=\columnwidth]{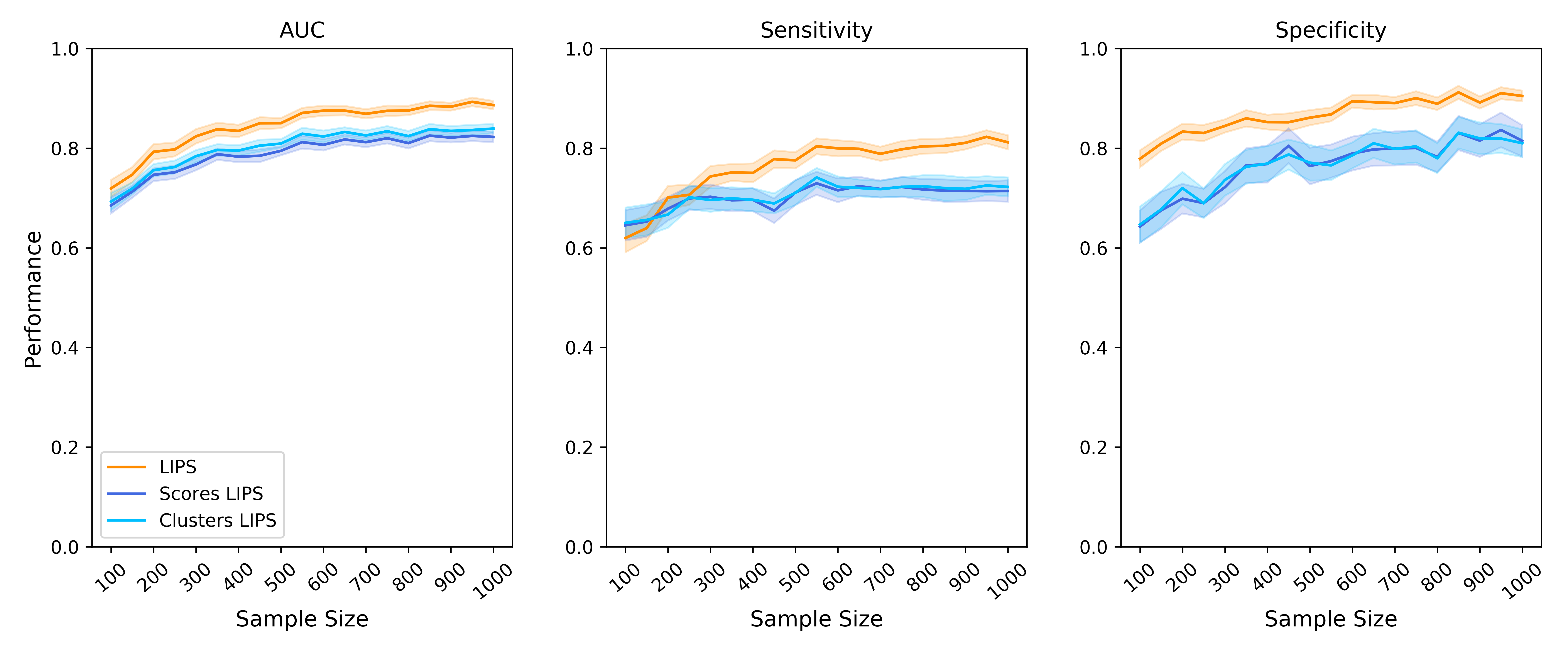}}
\caption{Performance of \texttt{LIPS}, \texttt{Scores LIPS} and \texttt{Clusters LIPS} for varying sample sizes.}
\label{fig:sample_size}
\end{figure}
Nowadays, real-life research settings present an extremely wide range of different scenarios when it comes to sample size. Huge data sets in some domains are opposed to extremely limited samples in others, above all the healthcare or medical research field. Therefore, any novel statistical approach that aims at finding broad application needs to be flexible to the different situations it might be applied to.

This experiment is meant to evaluate and discuss the performance of \texttt{LIPS} for extremely limited sample sizes. Indeed, as $n$ becomes larger, the only expected drawback of the proposed methodology regards computational time. It has already been discussed which safety measures were taken to limit the impact of extremely large datasets, therefore the focus here will be around the potential complexities in implementing \texttt{LIPS} on very few observations.

Indeed, mining item sets on one class only may easily induce low generalization capability of the identified interactions, if the sample is too small or poorly representative of the real class' distribution.

Moreover, decreasing minority class sample size may bias the structure (and reduce the number) of \textit{patterns} to evaluate for inclusion as interaction terms. Indeed, as the sample size gets smaller, fewer levels of each covariate may be represented in minority class examples.\\
With this experiment we wish to empirically demonstrate that thanks to the combined effect of the $supp_{min}$ parameter, of the OCL search, and the dissimilarity-based feature selection, \texttt{LIPS} can handle extremely limited sample sizes granting a good performance on the evaluation metrics. Indeed, searching on minority class examples with a low threshold on $supp_{min}$ allows the algorithm to identify a sufficiently large set of interactions despite working on very few observations. Moreover, the dissimilarity-based selection lowers the risk to overfit the small pool of observed data points and include in the final model interactions that might generalize better on the overall population.

For the experiment we trained \texttt{LIPS}, \texttt{Scores LIPS} and \texttt{Clusters LIPS} on datasets of varying sample size and $K=10$. For each sample size ($n \in \{100,1000\}$ with a step of 50 observations) we generated 50 training sets and test sets. In Figure \ref{fig:sample_size} we provide the obtained results in terms of mean and standard deviation on AUC, Sensitivity and Specificity. As the reader may notice, despite sample size gets significantly small and the number of terms included in the LR is rather limited (10 terms at most), the performance is not degrading significantly (lowest AUC score around 0.7 with the smallest sample of 100 observations only). Among the three, \texttt{LIPS} with ungrouped interactions performs significantly better than the other two versions. Again, this is not surprising as we are adding more terms to the LR without any information preprocessing. Nonetheless, the two other versions provide a solid performance as well, and might be useful in the case the number $K$ of interactions had to grow larger, or to satisfy specific interpretability requirements.

\subsection{On the robustness to Class Imbalance}
\begin{figure}
\centering
\makebox{\includegraphics[width=\columnwidth]{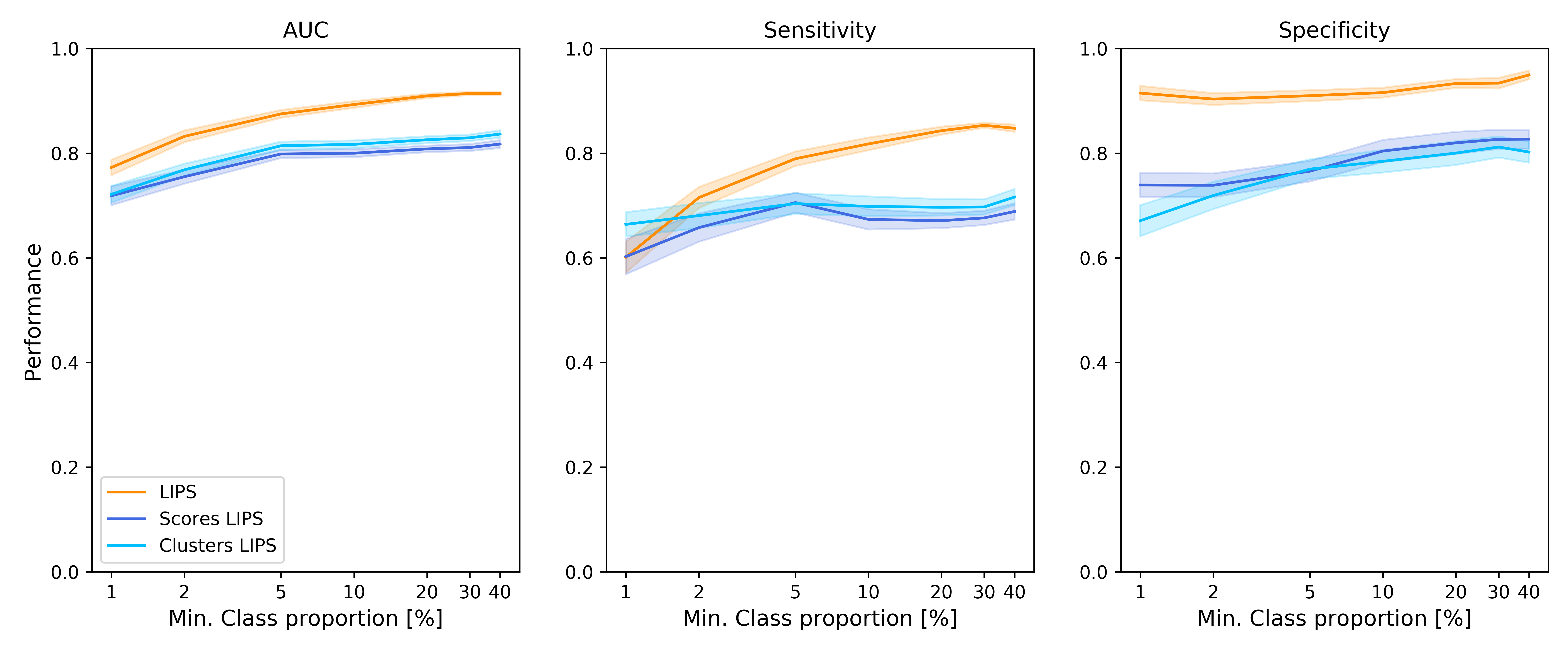}}
\caption{Performance of \texttt{LIPS}, \texttt{Scores LIPS} and \texttt{Clusters LIPS} for varying imbalance ratios. The percentages on the x-axis (reported in log-scale) represent the portion of minority class observations on the whole dataset.}
\label{fig:imbalance}
\end{figure}
The last relevant aspect to be discussed regarding the broad applicability of our proposed approach, is its capability to handle situations of strong class imbalance. As a matter of fact, the nature of the algorithm itself makes it almost unaffected by class imbalance ratios if the number of observations within minority class is sufficiently large. Indeed, the one-class-learning procedure reduces the effect of class imbalance, that would instead make the problem rather intractable in case the algorithm searched on the whole dataset together.

Let us consider an extreme case in which the itemset $i$ has empirical frequency $f_{i,1}\simeq 1$ among minority class and $f_{i,0}\simeq 0$ within majority observations. In what follows we write $n_{1}$ for the number of minority class observations and we let $p_{1}:=n_{1}/n$.
Then, in order for the algorithm to include $i$ (which is extremely relevant in discriminating between the two classes) in the list of interactions we should impose $supp_{min}<p_{1}$. In strongly imbalanced settings, $p_{1}$ can easily take values below 0.1. This would mean that including patterns specific to the positive class would force the imposition of extremely low $supp_{min}$ values, causing the list of potential interactions to explode. This would increase computational time, noise, and lower the probability of including strongly discriminative interactions in the model despite the high associated ORs.

Instead, the OCL approach of \texttt{LIPS} overcomes this limitation and identifies relevant interactions irrespectively of the imbalance ratio. For this experiment we generated 50 training and test set for each percentage of minority class observations, from $p_{1}:= n_{1}/n = 1\%$ to $p_{1} = 40\%$, with $n = 5,000$. Minimum support was set to 0.1 and $K=10$. In Figure \ref{fig:imbalance} it can be noticed how none of the presented metrics are strongly affected by the incresingly imbalanced setting of the classification problem. Of course, in this case the algorithm was dealing with a rather large $n$. However, as we already discussed the performance for small sample sizes, we were more interested in observing the effect of the imbalance ratio only. The three variants demonstrate a solid performance for extreme imbalancement as well, and (especially for traditional \texttt{LIPS}) basically reach and robustly keep their best performance from 5$\%$ on.

\section{Benchmark Experiments}
As discussed in Section \ref{related_works}, our proposition lies at the crossing of two independent research lines, building on concepts borrowed from the best of both.

In particular, the group of works dealing with \textit{significant pattern mining} (such as \cite{pellegrina2018efficient, sugiyama2019finding}) provides a set of potential alternatives for the identification of the list of interaction to be filtered on a \textit{dissimilarity} basis. As a first development of our methodology we decided to apply the most widely recognized item set mining algorithm, the Apriori algorithm. Nonetheless, future developments might include alternative pattern mining methods that might scale better, improve the efficiency or handle some data and problem-specific requirements more effectively. As a matter of fact, we see these methodologies more as potential additions to our algorithm, than actual competitors. Therefore, to compare the performance of our algorithm with a competing benchmark, we focused on the second line of research mentioned among related works. In particular, we chose the recent work of \cite{lim2015learning}, where the authors present Group-Lasso INTERaction-NET - \texttt{glinternet}, a state-of-the-art approach when it comes to selecting interactions for LR models.

For readability and comparability of results, there are a few considerations to be made regarding \texttt{glinternet}. The algorithm is indeed a regularization-based interaction learning method and as such it does not allow for a precise control over the number of terms included in the model, especially in case of categorical covariates. In facts, the open source software\footnote{\text{ } cran.r-project.org/web/packages/glinternet/index.html} allows the user to define the number of interactions ($n_{inter}$) to include in the model, and then the algorithm iteratively relaxes the regularization to get as close as possible to $n_{inter}$ interactions. The procedure stops when the algorithm reaches \textit{at least} $n_{iter}$, meaning that the number of interactions may be larger than required. Note that, \texttt{glinternet} includes main effects and interactions by enclosing the whole variables, with all their categorical levels. 

To compare \texttt{LIPS} with \texttt{glinternet} we applied both algorithms to simulated data and to a dataset from UCI Machine Learning Repository \citep{Dua:2019}, namely the Breast Cancer \footnote{\text{ }https://archive.ics.uci.edu/ml/datasets/breast+cancer} dataset.

\subsection{Benchmark comparison on simulated data}
To test \texttt{LIPS}'s performance against the benchmark, we run \texttt{glinternet} on 10 simulated training and test sets, generated as described in Section \ref{sim_dataset}. To compare results on the basis of how many terms were introduced in the LR as well, we imposed several different $n_{inter}$ values ($n_{inter}\in\{3,4,5,6,7,8,10\}$). Then, we collected the number of $\beta$ parameters associated with each level of the selected covariates to count precisely the number of terms in the model. Table \ref{tab:glinter1} reports the results of the experiment. Again, \texttt{LIPS} was trained with $K=10$ and $\threshold = 0.1$.

Our proposed algorithm performs notably better than the benchmark competitor on average, up until \texttt{glinternet} includes 10 interaction terms in the model. However, using 10 interactions translates in including on average 50 or 60 terms in the LR, as opposed to \texttt{LIPS} that is using 10 \textit{pattern}-terms only. This makes the resulting model way more interpretable, and the estimated $\beta$ parameters more reliable.

\begin{table}
\caption{\label{tab:glinter1}Results on the simulated dataset for \texttt{glinternet} (with varying number of interactions and consequent number of terms) and \texttt{LIPS} (last row).}
\resizebox{\textwidth}{!}{
\fbox{%
\begin{tabular}{*{6}{l}}
\textbf{Param} & \textbf{AUC} & \textbf{Sensitivity} & \textbf{Specificity} & \textbf{Mean N Vars}  & \textbf{Fitting Time {[}s{]}} \\ \hline
$n_{inter}=3$ & $0.852 \pm 0.016$ & $0.889 \pm 0.023$ & $0.781 \pm 0.039$  & 27 & $0.308 \pm 0.03$ \\
$n_{inter}=4$ & $0.881 \pm 0.011$ & $0.872 \pm 0.034$ & $0.844 \pm 0.039$  & 36 & $0.328 \pm 0.022$ \\
$n_{inter}=5$ & $0.888 \pm 0.015$ & $0.877 \pm 0.011$ & $0.863 \pm 0.037$  & 38.8 & $0.347 \pm 0.025$ \\
$n_{inter}=6$ & $0.896 \pm 0.019$ & $0.873 \pm 0.011$ & $0.890 \pm 0.051$ & 41.4 & $0.354 \pm 0.03$ \\
$n_{inter}=7$ & $0.902 \pm 0.021$ & $0.867 \pm 0.015$ & $0.930 \pm 0.059$  & 43.6 & $0.473 \pm 0.32$ \\
$n_{inter}=8$ & $0.921 \pm 0.011$ & $0.86 \pm 0.012$ & $0.976 \pm 0.02$  & 49.8 & $1.66 \pm 1.01$ \\
$n_{inter}=10$ & $0.923 \pm 0.007$ & $0.859 \pm 0.010$ & $ 0.984 \pm 0.003$  & 60.6 & $3.22 \pm 0.75$ \\ \hline
K = 10 & $0.916 \pm 0.010$ & $0.854 \pm 0.016$ & $ 0.949 \pm 0.038$  & 10 & $1.185 \pm 0.053$ \\
\end{tabular}}}
\end{table}

\subsection{Breast Cancer Case Study}
\label{sec:second_example}
For this last benchmarking experiment it was chosen a freely available dataset from UCI Machine Learning Repository, namely the Breast Cancer Dataset. 
This dataset describes oncological patients in terms of their age, menopause state and further features regarding both the tumor itself (size, malignity, breast location etc.) and its treatment.
Thus, instances are described by 9 categorical attributes in total, some of which are continuous binned into categories and some are nominal.
Patients are splitted in two classes: one grouping 201 instances with tumor recurrency, and the other including 85 women with no recurrency events.

What makes this dataset interesting for this application is the rather small sample size (typical of a real-life medical experimental setting), and the varying number of levels per feature, ranging from 2 to 11. 
In Table \ref{tab:breast} the results of the application of the three versions of \texttt{LIPS}, against \texttt{glinternet}, a LR model and a Lasso Regression both without interactions. Because of the small sample size, the cross-validation was performed by bootstrapping training and test set 10 times with splitting ratio $70/30$.
\newline
\newline
Our algorithms were all trained with $\threshold = 0.3$ and the additional use of $90\%$ confidence intervals for the odds-ratios; we used different values for $K$, as reported in Table \ref{tab:breast}. 
\texttt{LIPS} and its variants consistently attain the best performance in terms of AUC and Sensitivity, using the smallest number of terms in the model. Lasso Regression and \texttt{glinternet} perform comparably, however the latter includes in the model an eccessive and disproportionate number of terms. Indeed, as the number of terms grows, the performance of \texttt{glinternet} worsens.

These results suggest that the interaction effect in describing the outcome in this particular dataset is not extremely evident, nor necessary. Indeed, the absence of interaction terms is not strong enough to hinder the classification capability of the Lasso. However, the co-occurrence of some specific dummy variables in the minority class seems to capture a relevant part of the variability in the data, allowing \texttt{LIPS} and its variants to provide the best performance with a very limited pool of patterns.

\begin{table}
\caption{\label{tab:breast}Results on the Breast Cancer dataset for \texttt{LIPS} - (i) Individual interactions, (ii) Scores and (iii) Clusters, (iv) LR without interactions, (v) Lasso Regression and (vi) \texttt{glinternet} (with varying number of interactions and consequent number of terms).}
\resizebox{\textwidth}{!}{
\fbox{%
\begin{tabular}{*{6}{l}}
\textbf{Method} & \textbf{Parameters} & \textbf{AUC} & \textbf{Sensitivity} & \textbf{Specificity}  & \textbf{Mean N Terms}  \\ \hline

\texttt{LIPS} & K = 4 & \textbf{0.710 $\pm$ 0.062} & \textbf{0.696 $\pm$ 0.097} & $ 0.679 \pm 0.069$ & \textbf{4}  \\
\texttt{Scores LIPS} & K = 12 & \textbf{0.730 $\pm$ 0.051} & \textbf{0.742 $\pm$ 0.065} & $ 0.664 \pm 0.068$  & \textbf{2}  \\
\texttt{Clusters LIPS}  & K = 12 & \textbf{0.728 $\pm$ 0.053} & \textbf{0.742 $\pm$ 0.073} & $ 0.659 \pm 0.069$  & \textbf{2.7 $\pm$ 0.483}  \\ \hline
No Int. LR & - & $0.648 \pm 0.050$ & $0.662 \pm 0.090$ & $ 0.643 \pm 0.115$  & 51  \\
Lasso LR & $\lambda = 10^{-2}$ & $0.704 \pm 0.038$ & $0.650 \pm 0.069$ & $ 0.734 \pm 0.094$ &  $10.2 \pm 1.229$  \\ \hline
\texttt{glinternet} & $n_{inter}=2$ & $0.700 \pm 0.040$ & $0.611 \pm 0.152$ & $0.759 \pm 0.163$ &  51.2  \\
\texttt{glinternet} & $n_{inter}=3$ & $0.702 \pm 0.043$ & $0.596 \pm 0.157$ & $0.782 \pm 0.163$ &  54.3  \\
\texttt{glinternet} & $n_{inter}=4$ & $0.701 \pm 0.041$ & $0.557 \pm 0.169$ & $0.819 \pm 0.138$ &  69.9  \\
\texttt{glinternet} & $n_{inter}=5$ & $0.699 \pm 0.040$ & $0.580 \pm 0.142$ & $0.793 \pm 0.137$ & 104.5  \\
\texttt{glinternet} & $n_{inter}=6$ & $0.700 \pm 0.040$ & $0.635 \pm 0.179$ & $0.734 \pm 0.217$ & 132.7  \\
\texttt{glinternet} & $n_{inter}=8$ & $0.699 \pm 0.045$ & $0.603 \pm 0.150$  & $0.763 \pm 0.119$ & 194.8  \\
\texttt{glinternet} & $n_{inter}=13$ & $0.686 \pm 0.042$ & $0.611 \pm 0.165$ & $ 0.728 \pm 0.206$ & 327.4  \\ 
\end{tabular}}}
\end{table}

\section{Discussion and Conclusions}
\label{sec:discussion}
In this paper we presented \texttt{LIPS}, an high-order interaction learning algorithm via targeted pattern search, to select high-order interactions among categorical covariates and build a LR model. Together with \texttt{LIPS}, we proposed two alternative strategies to represent and include the selected interactions in the model, namely \texttt{Scores LIPS} and \texttt{Clusters LIPS}, to address the problem-specific needs of the research community that uses LR for inference and modeling.
The proposed strategies have been assessed through a wide set of experiments on both simulated and real data. In particular, as \texttt{LIPS} was designed specifically to address a broad range of real-life data analysis issues, we wish to conclude the paper by recalling and discussing the most notable pros of the algorithm.  

First and foremost, \texttt{LIPS} was designed to make LR models more powerful by introducing interaction terms between categorical covariates. However, to make the resulting model reliable for inference, a strict control over model dimensionality is needed.  To this specific aim, the option to choose the precise number of interaction terms to incude allows the user to tailor the LR fitting on the problem at hand, managing the risk of biased estimates and unreliable inference \citep{sur2019modern} induced by $p\gg n$ settings.
Controlling for the dimensionality of the model surely makes \texttt{LIPS} extremely effective in small sample size settings. Nonetheless, the algorithm is fast and scalable enough to be applied to research scenarios where both $p$ and $n$ are large. Indeed, the proposed method provides a useful feature selection technique, that reduces the computational cost by focusing its pattern search on one class only (\textit{targeted} search) without losing generalizability, and that returns lean and interpretable models in manageable running times.

Besides inference robustness, model dimensionality surely impacts interpretability as well. Most real-life application domains are willing to sacrifice a little on the performance side, to foster readability and explainability of results. \texttt{LIPS} addresses this aspect by providing an arbitrarily small and extremely predictive set of discriminative interaction terms, that are easy-to-read and interpret. We provided a clear example via simulation, where the selected patterns described precisely the areas where the two classes were to be sought for. Moreover, we introduced two alternatives to the principal algorithm (\texttt{Scores LIPS} and \texttt{Clusters LIPS}), whose aim is that of serving this need. Indeed, some potential applications of the two variants could be easily identified in the healthcare and lifescience domain, where $n$ is traditionally very small, $p$ may grow extremely large and interpretability is key. For instance, genome-wide association studies may require long lists of interactions to properly characterize a patient and introducing $K$ terms in the model may reduce both interpretability and reliability of results (as even $K$ may be larger than $n$). Furthermore, the \textit{Risk} and \textit{Protection Scores} of \texttt{Scores LIPS} provide a clear representation of the patterns and their role in profiling the population, while being an agile scoring system that can be easily accompanied by other explicative covariates.

Another point of paramount importance, is the ability of \texttt{LIPS} and its variants to manage even extremely imbalanced settings. We already highlighted how oftentimes minority class represents the most interesting class to study, while researchers need to make reliable inference on a very small sample of this critical population. \texttt{LIPS}, combining \textit{targeted} search with dissimilarity-based interaction selection, demonstrated to be a powerful tool to capture the underlying minority class distribution in a robust and generalizable manner, even in presence of very few observations.

Lastly, it is relevant to address the fact that \texttt{LIPS} and its variants focus their attention on the search for interactions among categorical covariates. This apparent limitation is actually grounded on a solid and multifaceted rationale. First of all, as previously mentioned, fully categorical data are getting more and more common in several scenarios, such as life sciences.  The tractability of these types of data within LR models with interactions is strongly affected by the exponential growth of the number of covariates to be considered as the number of levels grow. This same impact is less dramatic when dealing with continuous features. Nonetheless, \texttt{LIPS} was designed to solve this specific issue at best, while being a flexible and effective addition to a broader study framework with various data types as well. Indeed, once selected a restricted and powerful subset of $K$ interaction terms from the categorical subset of variables (absorbing a great portion of computational complexity), the user can easily accompany them with additional numerical covariates - and their interactions. These further terms might indeed be selected  - separately or jointly with the $K$ \texttt{LIPS}'s terms - via other traditional techniques, that could work on a reduced number of features w.r.t. using the whole original data.

We also compared the performance of the proposed algorithm with that of a state-of-the-art interaction selection method, \texttt{glinternet}, demonstrating our method's superior performance in terms of accuracy, interpretability, and significance of the resulting model, as \texttt{LIPS} and its variants obtained better results with notably less terms in the model.
\newline
\newline
In conclusion, with its \textit{targeted} approach, the introduction of a novel dissimilarity-based interaction selection, and its flexibility to be tailored around the specific user needs (i.e. number and order of interaction terms, interaction representation via Scores or Compatibility Clusters, confidence adjustment), \texttt{LIPS} results to be a novel and powerful approach to face the complexities of applying LR to many real-life research domains.\\
Future works may be devoted to improving the algorithm's efficiency by including more refined item set mining methods at its core and expanding the methodology to include numerical covariates in a noise-adverse manner.

\bigskip
\bigskip

\section*{Acknowledgements}

This project has received funding under the ERA PerMed Cofund program, grant agreement No ERAPERMED2018-44, RADprecise - Personalized radiotherapy: incorporating cellular response to irradiation in personalized treatment planning to minimize radiation toxicity.

\newpage

\appendix

\section*{Appendix}
In what follows, we assume to be given a dataset $\mathcal{D}=\{(x_{1,i}, \ldots, x_{p,i}, y_{i})\}_{i=1}^{n}$ of $n$ i.i.d. observations of $p$ categorical covariates $X_{1},\ldots, X_{p}$, resp. with $m_{1},\ldots,m_{p}$ levels, and an outcome $Y$. Again, we denote by $\mathcal{I}$ the set of all possible interaction terms for the $X_{i}$'s.

\section{Algorithmic Details}
\label{app:details}
\subsection{Support computation}
\label{app:support_comp}
After the mining phase, one is left with a candidate subset $\hat{\mathcal{I}}\subset\mathcal{I}$. In order to compute the corresponding odds-ratios, the datamatrix $\hat{\textbf{I}}$ has to be constructed; however, as the number of candidate patterns is tipically very large, evaluating each interaction term by its formula can be unnecessarily expensive. Parallel computing may surely help, but we can also exploit the particular structure of patterns to grant a faster computation.

Without loss of generality, let us rename the dummie variables $\{X_{1}^{(1)},\ldots,X_{p}^{(m_{p})}\}$ as $\{Z_{i}\}_{i=1}^{d}$, where  $d=\sum_{i=1}^{p}m_{i}$, and let $\textbf{Z}\in\mathds{R}^{n\times d}$ be the datamatrix of such dummies. For $|\hat{\mathcal{I}}|=L$ and $\hat{\mathcal{I}}=\{T_{k}\}_{k=1}^{L}$ we define the incompatibility matrix $\textbf{M}\in\mathds{R}^{d\times L}$ as (recall that here we allow dummies to be degenerate interactions)

$$\textbf{M}_{i,k}:=
\begin{cases}
1 & Z_{i}\perp T_{k}\\
0 & \text{otherwise}
\end{cases}\quad\quad\quad i=1,\ldots,d,\quad k=1,\ldots,L,$$
\smallskip

If we consider $\textbf{Z}, \textbf{M}$ and $\hat{\textbf{I}}$ as logical matrices (that is, matrices having entries in the boolean domain $\{0,1\}$), then, because each observation always attains exactly one level for each variable, it is very easy to see that

\begin{equation}
\label{eq:suppcomp}
\hat{\textbf{I}}=\neg\left(\textbf{Z}\cdot\textbf{M}\right),
\end{equation}

where the matrix product is intended in the boolean sense, whereas $\neg$ is the common notation for the logical negation operator (here applied entry-wise).
Let us briefly sketch the proof for \eqref{eq:suppcomp}. Consider the $j$th observation for the $k$th interaction, $i_{j,k}$. By definition of logic sum and product, the corresponding term on the right-hand side of \eqref{eq:suppcomp} is

$$\neg\left(\sum_{i=1}^{d}z_{j,i}m_{i,k}\right)=\neg\left(\exists i\in\{1,\ldots,d\}\;|\;z_{j,i} \land m_{i,k}\right)=$$

$$=\neg\left(\exists i\in\{1,\ldots,d\}\;|\;z_{j,i}\;\land\;Z_{j}\perp T_{k} \right) = \neg\left(\neg i_{j,k}\right) = i_{j,k},$$

in fact, the absence of a pattern in an observation is equivalent to the presence of (at least) a dummie it is incompatible with.

\bigskip
\bigskip

\subsection{Compatibility  clusters computation}
\label{app:clusters}
In Section \ref{sec:clusters}, we described the algorithm \texttt{Clusters LIPS}, which condensates the information coming from the interaction terms into fewer variables associated to certain compatibility clusters.
For more readability, let us recall the notation we used in that context. From the list of candidates $\hat{\mathcal{I}}$, we exploited the dissimilarity measure to extract $K$ patterns (with $K$ even). Half of those were selected from the risk interactions, resulting in the sublist $\hat{\mathcal{R}}_{K/2}$; similarly, the other half consisted of protective patterns, $\hat{\mathcal{P}}_{K/2}$.
Next, our algorithm requires the partitioning of these two lists (separately) into smaller groups of patterns so that: each group is formed by pairwise compatible patterns; different groups always have -at least- two incompatible patterns.
For the sake of simplicity, we address this problem only for the risk patterns.

\smallskip

\begin{figure}
\begin{center}
\makebox{\includegraphics[width=300pt]{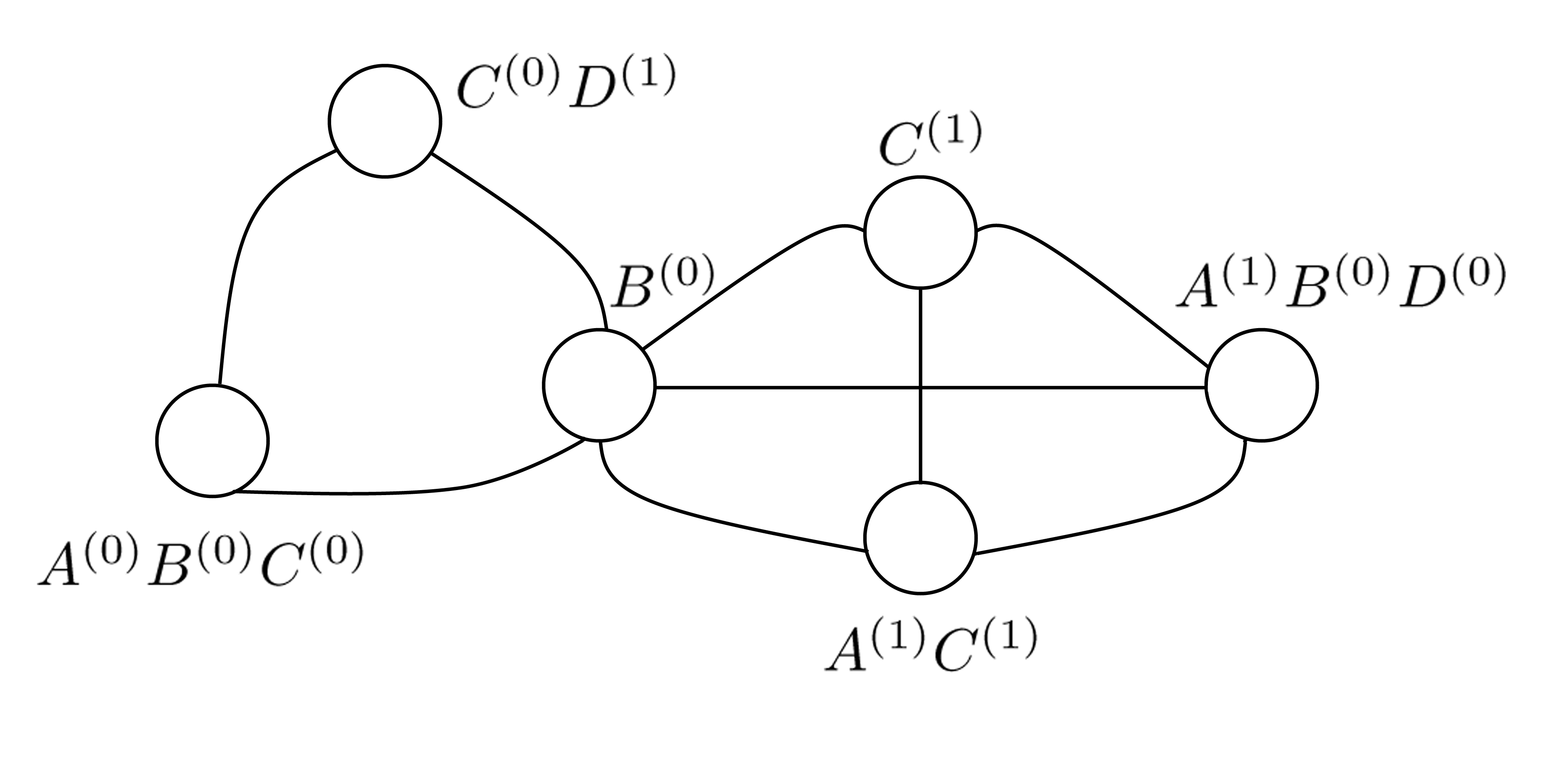}}
\caption{Example of a compatibility graph. Several interactions terms (two of which are degenerate) involving the dummies of four $\{0,1\}$-valued random variables, $A, B, C$ and $D$, define the nodes within the graph. Two nodes share a link if and only if the corresponding interactions are compatible. Our greedy decomposition breaks the graph into two clusters: the maximal clique $\{B^{(0)}, C^{(1)}, A^{(1)}C^{(1)}, A^{(1)}B^{(0)}D^{(0)}\}$ and the remaining pair $\{A^{(0)}B^{(0)}C^{(0)}, C^{(0)}D^{(1)}\}$.} 
\label{fig:cgraph}
\end{center}
\end{figure}

To determine a suitable partition of $\hat{\mathcal{R}}_{K/2}$ into compatibility clusters $\hat{\mathcal{R}}_{K/2}^{1},\ldots,\hat{\mathcal{R}}_{K/2}^{a}$, it can be convenient to consider the interactions as nodes of an undirected compatibility graph $\mathcal{G}(\hat{\mathcal{R}}_{K/2})$, where two interactions are linked together if and only if they are compatible. An example can be found in Figure \ref{fig:cgraph}. In that way, our original problem becomes equivalent to that of partitioning a graph into cliques, with the additional constraint that such cliques should not be further joinable into a larger ones. As foretold, in general the solution to such puzzle is not unique and many partitions are feasible; in view of dimensionality reduction, one may thus want to strengthen the constraints and search for a minimal cover. However, the problem of finding a minimal clique cover for a graph is known to be NP-hard \citep{karp}, so we partially avert this issue by opting for a greedy procedure. In short: iteratively, we find the maximal clique in $\mathcal{G}(\hat{\mathcal{R}}_{K/2})$, store its nodes in the new list $\hat{\mathcal{R}}_{K/2}^{j}$, and remove the clique from the graph; we keep this up until the graph becomes empty. 

The determination of the maximal clique can be done in several ways. For the implementation of \texttt{LIPS} Clusters, we actually relayed on the more general Born-Kerbosch algorithm \citep{kerbosch}, which in principle allows one to list \textit{all} the maximal cliques within a graph. This algorithm is known for being computationally expensive on large graphs, but that was not our case as we typically picked small values for $K$. Whenever large values of $K$ are needed, it may be convenient to relay on other maximal clique search algorithms \citep{cliques1, cliques3, cliques2, cliques4}.

\subsection{Further technicalities}
\label{app:tech}
Here we address a few last technical details that can be considered when implementing \texttt{LIPS} or one of its variants.

First of all, even by exploiting the most sophisticated routines certain datasets may still yield too high computational times in the first training phase. Aside from replacing $Apriori$ with other frequent itemsets mining algorithms, another possibility is that of reducing the interaction search only to those patterns that have a length smaller than a fixed $L>0$. This further constraint can be easily added as it does not impact on the subsequent steps of the algorithm. Fixing a maximal length $L$ can also be used to intentionally bound the order of the interactions, which may be of interest depending on the context.

Finally, another possible issue is that of finding interactions $T$ that along the dataset have an uncomputable rank, $\log|OR_{T}|$. This happens everytime the contingency table of $(T, Y)$ has at least an empty cell. To deal with such situations, several approaches can be adopted. For example one may either: 1) further investigate the anomaly and decide whether to discard the interaction or force its presence in the final list of patterns; 2) adopt an adjusted version of the OR, for instance by employing the Haldane-Anscombe correction \citep{or05}, which consists in preliminary increasing each value in the contigency table by 0.5.

\section{On the dissimilarity measure}
\label{app:math}
Here, we wish to point out a few theorical facts about the dissimilarity measure $d$ (\ref{eq:dissimilarity_measure}), introduced in Section \ref{sec:dissimilarity}. First of all, the proposed dissimilarity measure defines a so called \textit{semi-metric} over $\mathcal{I}$ \citep{semimetric}.
In fact, as the MCD of two interactions is always a subinteraction of them both, it is straightforward to see that $d$ satisfies:
\begin{itemize}
    \item[1)] $d(T, S)\ge0$ for all $T, S\in\mathcal{I}$ (positivity);
    \item[2)] $d(T, S)=0\iff T=S$ (identity of indiscernibles), 
    \newline in fact $d(T, S)=0$ if and only if either $|T|=|S|=0$ or $|\text{MCD}(T, S)|=|T|=|S|$, which is equivalent to $T=S$;
    \item[3)] $d(T, S)=d(S, T)$ (symmetry).
\end{itemize}

\smallskip

In general though, $d$ does not define a metric over $\mathcal{I}$, since the triangular inequality does not hold. As a counterexample consider the case of three binary variables $A, B$ and $C$. Then, for $T:=A^{(0)}B^{(0)}C^{(0)}$, $S = A^{(0)}B^{(0)}C^{(1)}$ and $Z = A^{(0)}B^{(0)}$, because $T\perp S$, one has

$$d(T, S)= 3 > 2 = d(T, Z) + d(Z, S).$$

\bigskip

\bibliography{biblio} 
\bibliographystyle{chicago}

\end{document}